\documentclass[runningheads]{llncs}

\usepackage[T1]{fontenc}
\usepackage{graphicx}
\usepackage{verbatim}
\usepackage{times}
\usepackage[numbers]{natbib}
\usepackage{multicol}
\usepackage[bookmarks=true]{hyperref}
\usepackage{units}
\usepackage{xfrac}
\usepackage{xcolor}

\begin{document}

\title{A Minimalist Controller for Autonomously Self-Aggregating Robotic Swarms: Enabling Compact Formations in Multitasking Scenarios}

\titlerunning{Minimalist Controller for Self-Aggregating in Multitasking}

\author{
Maria Eduarda Silva de Macedo\orcidID{0009-0003-8153-6359} \and 
Ana Paula Chiarelli de Souza\orcidID{0009-0007-5792-9549} \and 
Roberto Silvio Ubertino Rosso Jr.\orcidID{0000-0002-9691-8750} \and 
Yuri Kaszubowski Lopes\orcidID{0000-0002-4627-5590}
}
\authorrunning{M.E.S. de Macedo et al.}

\institute{Santa Catarina State University, Joinville SC 89219-710, Brazil}

\maketitle

\begin{abstract}
The deployment of simple emergent behaviors in swarm robotics has been well-rehearsed in the literature.
A recent study has shown how self-aggregation is possible in a multitask approach --- where multiple self-aggregation task instances occur concurrently in the same environment. 
The multitask approach poses new challenges, in special, how the dynamic of each group impacts the performance of others.
So far, the multitask self-aggregation of groups of robots suffers from generating a circular formation --- that is not fully compact --- or is not fully autonomous.
In this paper, we present a multitask self-aggregation where groups of homogeneous robots sort themselves into different compact clusters, relying solely on a line-of-sight sensor.
Our multitask self-aggregation behavior was able to scale well and achieve a compact formation.
We report scalability results from a series of simulation trials with different configurations in the number of groups and the number of robots per group.
We were able to improve the multitask self-aggregation behavior performance in terms of the compactness of the clusters, keeping the proportion of clustered robots found in other studies. 

\end{abstract}

\section{Introduction}

Self-aggregation is one of many emergent behaviors widely encountered in nature \cite{camazine2001} and applied in swarm robotics, as it serves as a prerequisite for other forms of cooperation \cite{dorigo2004}.
It is a behavior driven by decentralized control, where robots placed in an environment come together into a single, compact cluster as quickly as possible \cite{gauci2014DARS_aggregation}.

Self-aggregation has been achieved using reactive and fully recurrent neural network controllers \cite{gauci2014DARS_aggregation, gauci2014IJRR_aggregation}.
In \cite{gauci2014DARS_aggregation}, an evolutionary approach was employed to optimize the parameters of recurrent and reactive controllers, mapping the readings from the robots' line-of-sight sensor to the speeds of their differential wheels.
Out of 100 controllers obtained with each approach, compact formations were achieved in 99 and 19 cases for the recurrent and reactive approaches, respectively.
For the reactive controller, a grid search was also performed to fit the parameters, resulting in a compact formation.
In \cite{gauci2014IJRR_aggregation} it is proven that a long-range sensor is required for the autonomous self-aggregation controller.
However, \cite{gauci2014DARS_aggregation, gauci2014IJRR_aggregation} focus only on single-task self-aggregation.

In \cite{krischanski2023_IROS}, the autonomous self-aggregation of robots and clustering of objects using a line-of-sight sensor is generalized for a \emph{multitask} setup. 
This allows multiple instances of the same task to be performed concurrently at the same time and in the same environment without any external input.
However, in \cite{krischanski2023_IROS}, the multitasking self-aggregation results in an emergent behavior characterized by a non-compact and non-scalable circle formation.
In the context of segregation in swarm robotics, \cite{mitrano2019} achieved a similar circle formation with grid search.
In \cite{rezeck2021}, a compact formation was obtained using Gibbs Random Fields based on segregation and flocking behaviors. 
Their approach requires the relative position and velocity of neighboring robots (estimated with an external motion capture system).
In this paper, we perform a grid search to find parameters for multitask self-aggregation that produce compact clusters, relying solely on the robot's own sensors (an embedded camera). 


\section{Methodology}
\label{sec:methodology}

Based on the multitask generalization of the robot self-aggregation problem from \cite{krischanski2023_IROS}, we developed a controller that results in a compact formation.
The multitasking approach allows multiple instances of the same task, self-aggregation in this case, to be performed at the same time and environment.
Each task corresponds to a group of robots; we use different colors to identify the groups.

Consider the behaviors where $r \geq 1$ differential-wheeled robots scattered in a bounded rectangular environment cluster themselves, forming a single aggregate containing all $r$ robots \cite{gauci2014DARS_aggregation, gauci2014IJRR_aggregation}.
The generalized version of this problem (multitask) consists of dividing the $r$ robots into $g \geq 1$ groups \cite{krischanski2023_IROS}.
The $r$ robots must quickly sort themselves into $g$ aggregates, each containing all the robots from the same group.

In this study, we consider a simulated model of the e-puck robot \cite{epuck2009}.
Its diameter, height, and weight are approximately $\unit[7.4]{cm}$, $\unit[5.5]{cm}$, and $\unit[150]{g}$, respectively.
The robot
consists of several sensors and actuators, including a simulated camera located at its front, used in this study to implement a line-of-sight sensor. 
It has a pair of differential wheels capable of being independently set to move forward and backward at a simulated maximum speed of $\unit[12.8]{cm/s}$.

As in \cite{gauci2014DARS_aggregation, gauci2014IJRR_aggregation, krischanski2023_IROS}, each robot is equipped with a line-of-sight sensor, $I$, at its front.
The sensor indicates what is directly ahead of the robot: nothing (i.e., walls of the environment), another robot of the same group, or another robot from a different group.
We adopted $I=0$ to indicate nothing.
$I=1$ indicates another robot of the same group and 
$I=2$ indicates another robot from a different group.
Note that in the single-task case, there is no distinction regarding other robots being of the same group or another group.

As in \cite{gauci2014DARS_aggregation, gauci2014IJRR_aggregation, krischanski2023_IROS}, we implement a reactive control, which implies that the robots operate without memory, i.e., they do not have a current state nor store past readings.
Their actions must be based solely on instantaneous responses to the current sensor's reading $I$.

Therefore, the control structure is a map from the sensor's reading $I$ to a pair, $(\bar{v}_{\ell}, \bar{v}_{r}) \in \left[-1,1\right]^{2}$, of normalized angular velocities for the robot's left and right wheels, respectively.
Negative values mean backward movement and positive values mean forward movement.
Let $\bar{v}_{\ell,i}$ and $\bar{v}_{r,i}$ denote the normalized angular velocity of the left and right wheel, respectively, when $I=i$, that is, the parameters of the controller.
Our control structure's parameters are the tuple
$
    \bar{\mathbf{v}} = \left( 
    \bar{v}_{\ell,0}, \bar{v}_{r,0}, 
    \bar{v}_{\ell,1}, \bar{v}_{r,1}, 
    \bar{v}_{\ell,2}, \bar{v}_{r,2} 
    \right),
$
with   
$\bar{\mathbf{v}} \in \left[-1, 1\right]^{6}$.
The objective is to find the parameters that achieve the desired behavior.
This step is guided by a cost function, where lower values indicate a better adherence of the controller's parameters to the desired behavior.

As in \cite{krischanski2023_IROS}, the cost function of the controller $\bar{\mathbf{v}}$ is given by running simulations and computing the performance according to a predefined metric.
Considering $r$ robots divided into $g$ groups.
Let $G_k : k \in \{1, \ldots, g\}$ be the set of robots in group $k$.
Let $\mathbf{p}_i^{\left(t\right)}$ and $\mathbf{p}_j^{\left(t\right)}$ represent the positions of robots $i$ and $j$, respectively, at time $t$, with $i, j \in \{1, \ldots, r\}$.
As a metric, we take the sum of the furthest distance among robots of the same group at time $t$ as:

$ 
    d^{\left(t\right)} = \sum_{k=1}^{g} 
    max(\| \mathbf{p}_i^{\left(t\right)} - \mathbf{p}_j^{\left(t\right)} \| : i, j \in G_k \wedge i \neq j).
$
The performance of the controller in a single simulation run is given by the weighted sum of $d^{(t)}$ during $T$ time steps:
$
    U\left(\bar{\mathbf{v}}\right) = \sum \limits_{t=1}^{T} t\,d^{\left(t\right)}.
$
Following \cite{gauci2014AAMAS_objectclustering, gauci2014DARS_aggregation, gauci2014IJRR_aggregation, krischanski2023_IROS}, the minimization of $U\left(\bar{\mathbf{v}}\right)$ seeks a low dispersion of elements at the end of the time interval and a fast reduction of the dispersion.
It achieves that by setting the weight of $d^{\left(t\right)}$ proportionally to the time $t$.
At the start of the run, the weight is low, and it gradually increases as the run approaches its end.

This weighted sum has two important characteristics, as the weight increases linearly with time, giving more importance to later measurements of $d^{(t)}$.
First, it heavily penalizes situations that do not converge to a compact formation or fail to remain steady. 
Second, it incentivizes fast convergence.

We employ the open-source Enki simulation platform \cite{Magnenat2005}.
Enki is a robot simulator written in C++ capable of handling the kinematics and the dynamics of rigid bodies in two dimensions (2D).
It has built-in models of a few robotics platforms, including the e-puck.
We adopt a control cycle of $\unit[0.1]{s}$ and set the physics update rate at $10$ times per control cycle (i.e. $100$ times per second).

\section{Controller Synthesis}
\label{sec:synthesis}

The control structure has six parameters ranging from $-1.0$ to $1.0$, where $-1.0$ means full reverse speed, $0$ means stopped, and $1.0$ means full forward speed. 
Our objective is to perform a grid search over the space of possible control parameter combinations in steps of size 0.1.
Each combination of parameters would be subjected to several simulation runs, with different initial positions in terms of the robot's position and orientation. 
However, this would represent a huge computational burden.


To alleviate the computational burden of executing a grid search over all six parameters, 
we keep the first four parameters of the controller unchanged, using the values found by
\cite{gauci2014IJRR_aggregation}: 
$\bar{\mathbf{v}} = (-0.7, -1.0, 1.0, -1.0, \bar{v}_{\ell,2}, \bar{v}_{r,2})$.
That is, we employ the same parameter for the single-task case when the robot is seeing nothing ($I=0$) or seeing another robot from the same group ($I=1$).
For the last two parameters ($\bar{v}_{\ell,2}$ and $\bar{v}_{r,2}$) of the controller, i.e., when the robot sees another robot from a different group, we perform a grid search over the entire space of possible combinations.
We used a resolution of 21 settings per parameter, with each parameter taking values in $\{ -1.0, -0.9, \ldots, -0.1, 0, 0.1, \ldots, 0.9, 1.0 \}$.

For each combination of the two parameters ($\bar{v}_{\ell,2}$ and $\bar{v}_{r,2}$) we performed $30$ simulations runs; resulting in $13230$ runs.
Each run was performed with $g=3$ groups for $\unit[2400]{s}$ on $r=75$ robots ($\sfrac{r}{g} = 25$ per group).
In each run, robots were randomly uniformly distributed in a virtual squared environment of sides $\unit[450]{cm}$ and their initial orientations were randomly uniformly distributed in the interval $[-\pi, \pi]$.

The desired end location can be guided by the presence of reference bollards, with one bollard assigned to each group. 
These bollards are fixed in place and share the same color as the robots in their respective group. 
The robots' sensors detect the bollards as if they were robots from their own group.
That is, robots sense the reference bollard as $I=1$.
Reference bollards were evenly distributed over a concentric circle of radius $\unit[40]{\%}$ of the length of the environment.


\setlength{\belowcaptionskip}{-15pt}
\begin{figure}[t]
    \centering
    \includegraphics[width=0.52\textwidth]{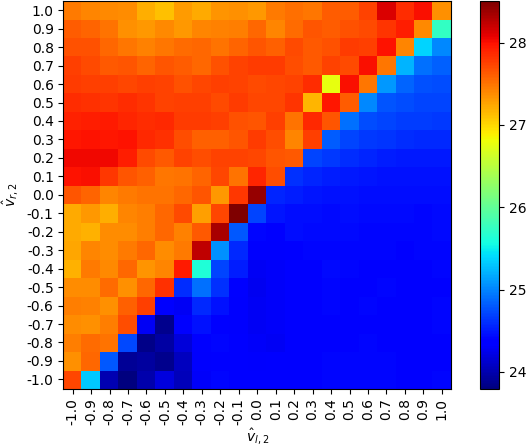}
    \vspace{-0.3cm}
    \caption{Heatmap showing the grid search results for the two parameters ($\bar{v}_{\ell,2}$ and $\bar{v}_{r,2}$)
    The color gradient, ranging from blue (lower) to red (higher), represents the natural logarithm of the average cost function across $30$ simulation runs for each parameter combination.
    Darker blue areas indicate lower (better) cost values, while red areas represent higher costs. 
    }
    \label{fig:gridsearch}
\end{figure}

Figure \ref{fig:gridsearch} shows a heatmap of the grid search results for the two parameters $\bar{v}_{\ell,2}$ and $\bar{v}_{r,2}$.
The result for each parameter configuration represents the average cost function ($U\left(\bar{\mathbf{v}}\right)$) across all $30$ simulation runs.
A distinct cut-off relationship is observed: better control performance (blue area) is obtained when the left speed (x-axis) is strictly greater than the right speed (y-axis) when the line-of-sight sensor is $I=2$ (i.e., $\bar{v}_{\ell,2} > \bar{v}_{r,2}$). 
The best parameters found were $(\bar{v}_{\ell,2} \bar{v}_{r,2}) = (-0.7, -1.0)$, resulting in the controller configuration:
$\bar{\mathbf{v}} = (-0.7, -1.0, 1.0, -1.0, -0.7, -1.0)$.

Note that this indicates that the best controller is to actuate the same speeds for the robot's wheels in both cases: when the line-of-sight sensor detects nothing ($I=0$) or a robot from another group ($I=2$).

In \cite{krischanski2023_IROS}, the core behavior involves scanning the environment by making sharp forward turns until a robot from the same group is spotted. 
When another robot from the same group is detected, the robot moves straight towards its direction.
If the robot stops sensing another robot from the same group, it returns to the scanning motion.

Our core behavior consists of performing a smooth backward clockwise turn until a robot from the same group is spotted.
Once another robot from the same group is detected, the robot rotates clockwise at full speed.
If the robot stops sensing another robot from the same group, it returns to the backward clockwise turn motion.

\section{Validation and Experimental Scalability}
\label{sec:validation}

We conducted systematical trials for the validation and to investigate the scalability in terms of the number of groups and robots per group.
We tested several configurations 
with $g=\{3, 5\}$ groups, 
$\sfrac{r}{g} = \{10, 15, 20, 25, 30\}$ robots per group (up to $150$ in total), and with and without reference bollards.
We conducted $10$ trials lasting $\unit[2400]{s}$ for each configuration.
Each trial was performed with different initial configurations (position and orientation) of the robots.

We employ two metrics: dispersion and proportion of clustered robots.
A low dispersion in each group indicates that the objective of forming compact groups of robots was achieved.
The second moment \cite{Graham:1990:a} is a measure of the dispersion of elements.
We use a group-based second moment of the robots to their group's centroid as a measure of their dispersion at each discrete time $t$. 
Let the centroid of the $n_k$ robots from group $k$ at discrete time $t$ be given by:
$
\mathbf{\bar{p}}^{\left(t\right)}_{k} = \frac{1}{n_k} \sum_{i=1}^{r} \mathbf{p}_i^{\left(t\right)} : i \in G_k.
$
Note that:
$
r = \sum_{k=1}^{g} n_k.
$

Let ${r_o}$ be the radius of the robots.
The group-based second moment at discrete time $t$ is given by:
$
    u^{\left(t\right)} = \frac{1}{4r_o^2} \sum \limits_{i=1}^{r} ||\mathbf{p}_i^{\left(t\right)} - \mathbf{\bar{p}}^{\left(t\right)}_{k}||^2 : i \in G_k.
    \label{eq:SecondMoment}
$
The proportion of clustered robots of the group $k$ indicates how many robots are already clustered together.
This is defined by the ratio between the largest graph of connected robots in the group, $lc_k$, and the total number of robots in each group, $n_k$; and is given by:
$
    pc_k = \frac{lc_k}{n_k}.
$
Two robots, $i, j$ of the same group, $k$ (i.e., $i, j \in G_k$), are considered connected if another robot cannot fit in between them.
That means that their distance (between centers) is smaller than four times their radius (i.e., $\| \mathbf{p}_i^{\left(t\right)} - \mathbf{p}_j^{\left(t\right)} \| < 4r_{o}$). The proportion of clustered robots of all groups is given by:
$
    pc = \frac{1}{r} \sum\limits_{k=1}^{g} lc_k.
$


In both metrics (dispersion and proportion of clustered robots) there is a noticeable convergence period followed by a steady state.
That means, that after clustering together, the obtained emergent behavior is capable of keeping the compact formation.

In terms of the dynamics of dispersion over time, comparing experiments with $g=3$ and $g=5$, both the cumulative dispersion and the time to achieve steady state increase slightly across all setups (with/without bollards and varying numbers of robots per group, $\sfrac{r}{g}$). 
The cumulative dispersion and the time to achieve a steady state also increase slightly as the number of robots per group rises. 
This is expected, as larger groups pose a greater challenge. 
Nonetheless, the additional time required to reach a steady state represents only a fraction of the increase in problem size, in terms of the number of robots.

When comparing the dispersion in cases with the bollard against those without it, it is noticeable that adding the reference bollard increases both cumulative dispersion and the time to achieve a steady state. 
This increase is almost unnoticeable with $g=3$, but more prominent with $g=5$. 
Traveling to the reference bollard location is an additional step, requiring extra time. 

In terms of the proportion of clustered robots, the experiments with three groups ($g=3$) had a good performance, especially in the presence of the reference bollard.
When considering five groups ($g=5$), the reference bollards greatly improve the proportion of clustered robots.
Indeed, due to the bounded arena size, occlusion plays a delaying factor in achieving the full connection of robots, and the reference bollards mitigate the negative effect of occlusion.
Even though occlusion still plays a role in our obtained behavior, it represents a significant improvement over the results reported in \cite{krischanski2023_IROS}.
Considering $\sfrac{r}{g} = 30$ and the use of the bollard, \cite{krischanski2023_IROS} reports the proportion of clustered robots at the $\unit[1800]{s}$ mark only at $40.22$\% and $38.53$\% for $g=3$ and $g=5$, respectively; in our obtained behavior, at the same mark, we have $100.00$\% and $99.33$\% for $g=3$ and $g=5$, respectively. 

\begin{figure}[t]
    \centering
    \begin{tabular}{cccccc}
        \includegraphics[width=0.156\textwidth]{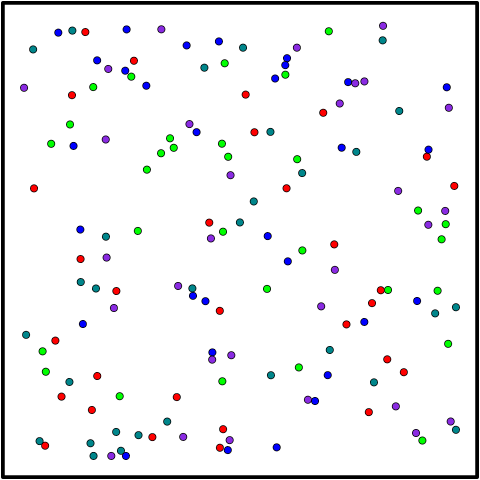} &
        \includegraphics[width=0.156\textwidth]{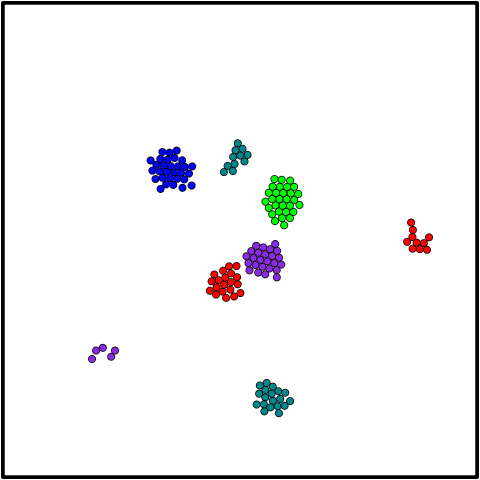} &
        \includegraphics[width=0.156\textwidth]{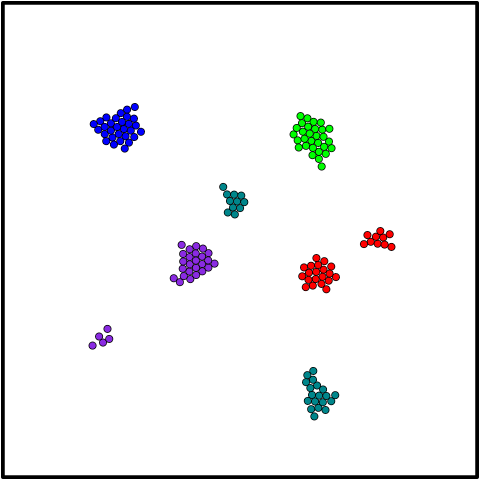} &
        \includegraphics[width=0.156\textwidth]{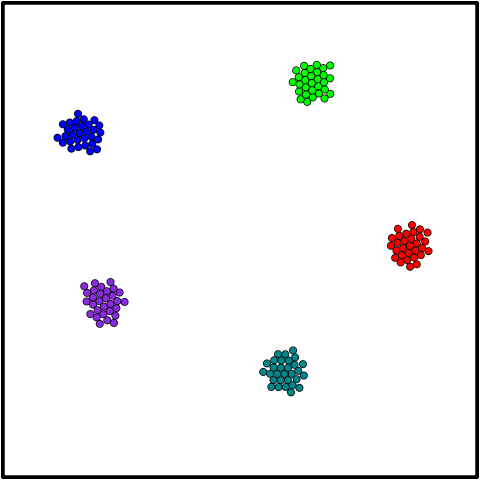} &
        \includegraphics[width=0.156\textwidth]{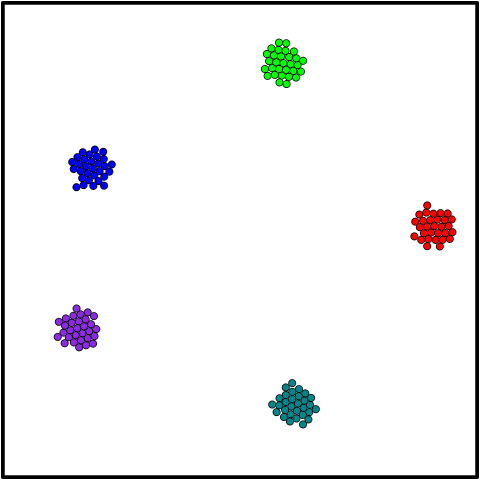} &
        \includegraphics[width=0.156\textwidth]{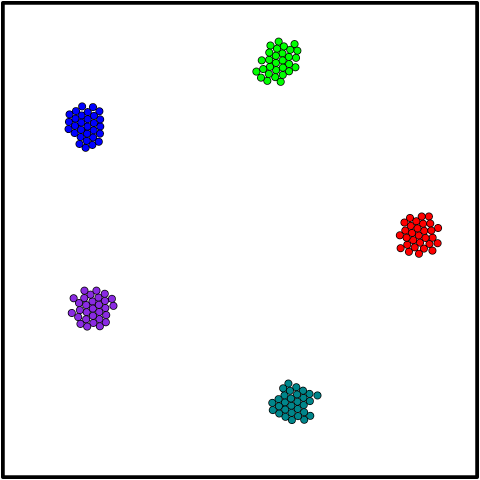} \\
        
        initial config. & $300$ seconds & $600$ seconds & $900$ seconds & $1200$ seconds & $2400$ seconds \\ 
    \end{tabular}
    \vspace{-0.2cm}
    \caption{
    A series of snapshots from an experimental trials where
    $r=150$ robots divided into $g=5$ groups of $\sfrac{r}{g} = 30$ robots per group perform the best controller for the multitask self-aggregation \underline{with} the presence of bollards.
    }
    \label{fig:snapshots}
\end{figure}

Figure \ref{fig:snapshots} presents a series of snapshots from one of the trials with $g=5$ groups and $\sfrac{r}{g} = 30$ robots per group, illustrating the compact formation behavior obtained.
The use of reference bollards influenced the cluster distribution around the center of the arena, within the radius defined by the reference bollards' distribution.
Without the reference bollards, all clusters tend to form near the center of the arena. 
When the reference bollards are not used, the behavior notably enables two clusters of robots of the same type (color) to eventually merge while navigating around other clusters.

\setlength{\belowcaptionskip}{-15pt}

\section{Conclusions}
\label{sec:conclusions}

In this paper, we performed a grid search to obtain an emergent behavior that achieves multitask self-aggregation with a compact formation.
Although the state of the art converges faster to its steady state, our behavior possesses better scalability performance in terms of the proportion of clustered robots after achieving the steady state, as well as overall better dispersion (compactness).
In future work, we will present an extended analysis of the behavior and propose a combined strategy that alternates between circular and compact formations to achieve faster convergence to a steady state, characterized by low dispersion and a high proportion of clustered robots.

\begin{credits}

\subsubsection{\ackname} FAPESC (2024TR1104, 2024TR2555) and Finep 1320/22.


\end{credits}

%
%
%
%
\bibliographystyle{splncs04}
\bibliography{references}

\end{document}